%% file: tae_workshop_arxiv.tex
\documentclass{article}

\PassOptionsToPackage{numbers,sort&compress}{natbib}
\usepackage[preprint]{neurips_2026}

\usepackage[utf8]{inputenc}
\usepackage[T1]{fontenc}
\usepackage{url}
\usepackage{booktabs}
\usepackage{float}
\usepackage{amsmath,amssymb,amsfonts,amsthm}
\usepackage{nicefrac}
\usepackage{microtype}
\usepackage{xcolor}
\usepackage{graphicx}
\usepackage{algpseudocode}
\floatstyle{ruled}
\newfloat{algorithm}{tbp}{loa}
\floatname{algorithm}{Algorithm}
\usepackage{enumitem}

\newtheoremstyle{thmbody}{3pt}{3pt}{\itshape}{}{\bfseries}{.}{.5em}{}
\theoremstyle{thmbody}
\newtheorem{theorem}{Theorem}[section]
\newtheorem{proposition}[theorem]{Proposition}
\newtheorem{corollary}[theorem]{Corollary}
\newtheorem{lemma}[theorem]{Lemma}
\newtheorem{assumption}[theorem]{Assumption}
\theoremstyle{remark}
\newtheorem{remark}[theorem]{Remark}

\newcommand{\Yref}{Y^{\mathrm{ref}}}
\newcommand{\muref}{\mu^{\mathrm{ref}}}
\newcommand{\Bsel}{B}
\newcommand{\Rref}{L}
\newcommand{\piref}{\pi^{\mathrm{ref}}}
\newcommand{\pidel}{\pi_{\delta}}

\newcommand{\Gmeas}{G_{\mathrm{meas}}}
\newcommand{\se}[1]{{\scriptstyle\pm#1}}


\title{Which LLM for Which Work?\\Budgeted Model Allocation under Uncertain Evaluation}

\author{Hamed Khosravi\\
  H. Milton Stewart School of\\
  Industrial and Systems Engineering\\
  Georgia Institute of Technology\\
  Atlanta, GA 30332\\
  \texttt{hkhosravi7@gatech.edu}
  \And
  Xiaoming Huo\\
  H. Milton Stewart School of\\
  Industrial and Systems Engineering\\
  Georgia Institute of Technology\\
  Atlanta, GA 30332\\
  \texttt{huo@gatech.edu}}

\begin{document}
\maketitle

\begin{abstract}
\looseness=-1 A company with a fixed artificial intelligence (AI) budget must decide which large
language model (LLM) handles each recurring workload. It knows each workload's traffic and each
model's price. What it lacks is the \emph{quality table}, how well each model performs on each
workload. Given that table, the decision is a multiple-choice knapsack problem and is routine to solve,
so estimating it is the difficulty, and that estimation fails in two ways. Models are rarely
compared on the same work, and the recorded score is usually a proxy rather than the outcome the
company values. Causal and off-policy methods repair the first but condition on the second, while
evaluator-validation methods estimate the second but stop short of the decision. Worse, buying more
re-evaluation cannot settle the second: randomization governs which requests are scored, not how a
score is produced, so the table stays uncertain however much evaluation is purchased. Yet the
deployment decision may still be determined even when the table is not. We therefore ask whether one assignment stays
optimal across every quality table consistent with the evidence. For the fixed-budget
problem, this admits an exact two-solve certificate: solve once at the estimated table and once at a
least-favourable table. Agreement certifies the assignment; disagreement identifies the
model--workload pairs where further evidence can matter. We propose CASE (causal active sequential
experimentation), which targets evaluation to those pairs and repeats the test as evidence
accumulates. On a production log, the measurement failure is the larger of the two: correcting
assignment exactly still leaves most of the loss, and randomized re-evaluation does not remove it. In our experiments, the available evidence often does not determine which model should be assigned
to each workload. On paid software tasks, better information about model quality yields more
savings than further optimization of the model-to-workload assignment using the same quality
estimates.
\end{abstract}

\section{Introduction}\label{sec:intro}
\looseness=-1 Consider a company with a fixed monthly budget for language models, whose engineers
write code, support team answers tickets, analysts summarize documents, and sales team drafts
outreach. Prices differ substantially and the same model need not be best for every kind of work, so
the company must decide which model serves which workload, and revisit it as prices and its mix of
work change. Such decisions govern much of the \$12.5 billion of annual enterprise spending on
foundation models \citep{menlo2025enterprise}.

\looseness=-1 Stated formally the problem is standard. Group the work into $J$ workloads and the
options into $K$ candidate models, and let the \emph{quality table} $\muref$ hold, in entry
$\muref_{kj}$, the average over all of workload $j$ of the outcome that model $k$ would deliver. We call
that per-request outcome the \emph{designated outcome}, the quantity the operator has chosen to
value and that the recorded score only proxies; its workload average is the cell's \emph{target
quality}. Given that
table, choosing one model per workload under a shared budget is a multiple-choice knapsack problem
\citep{sinha1979mckp,szkaliczki2025mckp}, and standard solvers handle far larger instances. The
allocation is not the difficulty. Estimating $\muref$ is, and the way that estimation fails is what
separates this problem from a knapsack problem whose coefficients are known.

\paragraph{Why $\muref$ is hard to estimate.} \looseness=-1 The table must be read off
production logs, automatic judges and user feedback, and two features of that data obstruct it.
Assignment is not independent of the outcome, since a default, a habit or an automated selection
system decided which model received a request
\citep{feng2025graphrouter,zhang2025routerr1,su2026cprouter,zheng2026disrouter}, so a cheaper model
can look competitive merely because it received easier requests. And the recorded score is a proxy,
since an LLM judge, a task grader or a click can depend systematically on answer length, answer
position, or which model produced the response
\citep{zheng2023judging,wang2024notfair,panickssery2024selfpreference,gu2025judgesurvey,dorner2025limits}.
These are failures of \emph{assignment validity} and of \emph{measurement validity}, both biases
rather than sampling error, and they respond differently to spending: re-evaluating a random sample
of requests removes the first and leaves the second, because randomization governs which requests
are scored and not how a score is produced.

\paragraph{Our approach: settle the decision, not the table.} \looseness=-1 The
operator does not need the quality table, only the decision it induces, so we ask not how to
estimate the table but whether \emph{the allocation is the same at every table the evidence admits}.
Call the decision \emph{settled} when it is, pinned down by the evidence in hand however wide that
set may be. Deciding this appears to need a continuum of solves. It does not: for the budgeted rule
one least favourable table dominates the set, so a second solve of the same program answers it
exactly, and the decomposable rule has an exact workload-wise analogue (App.~\ref{app:sat}). When the two solves disagree, the pairs on which they differ
are the ones worth paying to measure, and \S\ref{sec:protocol} spends a fixed evidence budget there.
We call the method \textbf{CASE} (causal active sequential experimentation): \emph{causal} because
the evidence it buys is a randomized re-evaluation rather than more of the same log, \emph{active}
because it buys only where the two solves disagree, and \emph{sequential} because the test is re-run
after each purchase.

\paragraph{How this compares with existing approaches.} \looseness=-1 Three lines of work each
supply part of what this problem needs. Optimization under uncertainty starts from the coefficients,
or a set of possible values for them, and finds the best decision
\citep{sinha1979mckp,szkaliczki2025mckp,kuhn2025dro,diwan2026explorable}, but begins where we
cannot, since that set is what we must derive from data. Causal inference and off-policy evaluation recover what each model would have scored
on the whole workload when assignment is as good as random given observed features
\citep{tsiourvas2025causal,zhang2025metarouter,gao2026budgeted,frauen2026causal}, but take the
recorded score at face value, so an evaluator that quietly prefers one model passes undetected,
looking exactly like that model being better (Cor.~\ref{cor:blind}). Work on evaluator validity
measures how far a score departs from human judgement
\citep{bean2025construct,wallach2025measurement,boyeau2025autoeval,angelopoulos2023ppi} without
asking whether a gap of that size would change a deployment. Each addresses one half. Our problem
requires both at once (App.~\ref{app:related}). Six corpora test these claims in
\S\ref{sec:results}.
\section{Problem formulation}\label{sec:setup}\label{sec:rules}\label{sec:criterion}\label{sec:dgp}\label{sec:target}\label{sec:obs}
\looseness=-1 An operator must put one model on each workload. Prices, traffic shares and the
serving budget are known, and the only unknown input is a table of quality coefficients.
App.~\ref{app:notation} lists every symbol.

\looseness=-1 Two allocation rules are common in practice, both taking a quality table and
returning one model per workload. Workloads are indexed $j=1,\dots,J$ and models $k=1,\dots,K$. Workload $j$ carries
traffic share $w_j$ with $\sum_{j=1}^{J}w_j=1$, model $k$ costs $p_{kj}$ per request on it, the
operator serves $N$ requests per period under a serving budget $\Gamma$, and $\mathcal K_j$
collects the models supported on workload $j$, all known. We state the rules for a
generic table $\mu$, since the difficulty throughout is that the operator must supply a table that
is not $\muref$.

\begin{figure}[H]\centering
\includegraphics[width=0.84\textwidth]{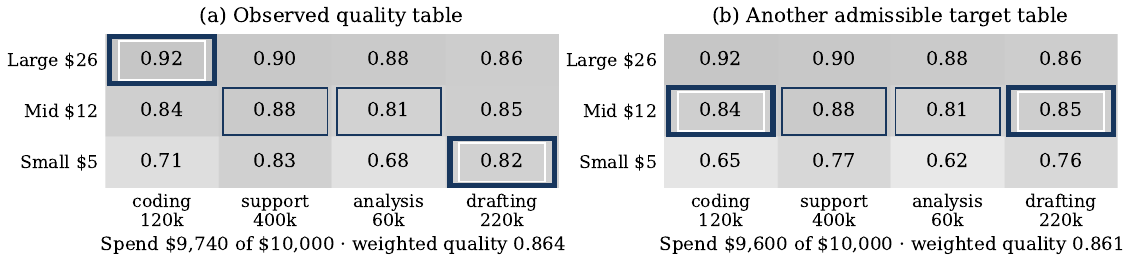}
\caption{\textbf{Illustrative budgeted allocation}, not an empirical result. Rows are models with
per-thousand-request prices, columns are workloads with monthly volumes, cells report quality, and
outlined cells show the allocation, the heavier outline marking the two decisions that differ. The
right panel changes only the Small model's target quality, yet coding and drafting both move.}
\label{fig:alloc}
\end{figure}

\paragraph{Fix the spend, maximize quality.} With a binary $x_{kj}\in\{0,1\}$ selecting model $k$
for workload $j$,
\begin{equation}\label{eq:mckp}
  X^\star[\mu]=\arg\max_{x}\sum_{j}\sum_{k\in\mathcal K_j}w_j\mu_{kj}x_{kj}
  \ \ \text{s.t.}\ \
  N\sum_{j}\sum_{k\in\mathcal K_j}w_jp_{kj}x_{kj}\le\Gamma,
  \ \ \sum_{k\in\mathcal K_j}x_{kj}=1\ \forall j.
\end{equation}
\looseness=-1 $X^\star[\mu]$ is the set of best affordable allocations for the table $\mu$, more
than one when they tie. The last constraint picks one model per workload, written $x(j)$. The single budget
constraint \emph{couples} the workloads, so the choices cannot be made separately
(Fig.~\ref{fig:alloc} in App.~\ref{app:notation}).

\paragraph{Fix the quality, minimize the spend.} With a tolerance $\delta\ge0$,
\begin{equation}\label{eq:sat}
  \pidel[\mu](j)=\arg\min_{k\in\mathcal K_j}
  \bigl\{\,p_{kj}\;\bigm|\;\mu_{kj}\ge\max_{k'\in\mathcal K_j}\mu_{k'j}-\delta\,\bigr\}.
\end{equation}
\looseness=-1 Call a model \emph{admissible} in workload $j$ when it is within $\delta$ of that
workload's best, so \eqref{eq:sat} buys the cheapest admissible model. It carries no budget
constraint, so unlike \eqref{eq:mckp} it \emph{decomposes}, and workload $j$ is settled by column
$j$ of $\mu$ alone. Neither rule is a special case of the other.

\paragraph{The decision map.} \looseness=-1 Both rules are maps from a quality table to an
allocation, $\Psi\colon\mu\mapsto x$. Both are written with an $\arg$, set-valued when two allocations tie. Ties are exactly the
boundaries this section is about, so we break them by a fixed total order, cheapest first and then
lexicographic, making $\Psi$ single-valued.

\looseness=-1 Statements using nothing beyond the map are made for $\Psi$; Prop.~\ref{prop:robust}
turns on the budget coupling of \eqref{eq:mckp} and is stated for that rule alone. Because
\eqref{eq:sat} decomposes, settlement there is checked column by column, by asking whether the
cheapest admissible model in a workload stays the same over that column's interval, and
App.~\ref{app:sat} gives the closed form.

\looseness=-1 $\Psi$ is \textbf{piecewise constant}. Both rules read only which cells beat which and
by how much, so a whole neighbourhood of tables returns the same allocation. Write $\tau(\mu)$ for the
\emph{stability radius}, the largest $t$ with $\Psi(\mu')=\Psi(\mu)$ for all
$\|\mu'-\mu\|_\infty\le t$. Error below $\tau$ is free, so the operator never needed $\muref$, only
the region it falls in. Each allocation $x$ owns the set of tables that produce it,
$\Sigma_x=\{\mu:\Psi(\mu)=x\}$, and these regions partition the space of quality tables. Their shape is the one place the rules
differ. Under \eqref{eq:mckp} the objective is linear in $\mu$, so $\Sigma_x$ intersects the
half-spaces on which $x$ beats each budget-feasible rival and a boundary is a hyperplane separating
two whole allocations; under \eqref{eq:sat} the rule decomposes, so $\Sigma_x$ is a product across
workloads and a boundary is a threshold on one workload's quality gap. Coupling makes the first case
global, so crossing one hyperplane can reassign several workloads at once
(Fig.~\ref{fig:alloc} in App.~\ref{app:notation}, where changing one model's quality moves two
assignments).

\paragraph{Which errors move the decision, and which do not.} \looseness=-1 Error of a given size is
not equally dangerous in every direction, and both rules agree on which directions matter.

\begin{lemma}[Columns are free, rows are not]\label{lem:shift}
Let $\Psi$ be either rule. Adding a constant $c_j$ to every entry of column $j$, one constant
per column, leaves the allocation unchanged. Adding a constant $c_k$ to every entry of row $k$ can
change it.
\end{lemma}

\noindent\looseness=-1 Under \eqref{eq:mckp} a column shift moves every allocation's objective by
the same $\sum_jw_jc_j$, since $c_j$ does not depend on $k$ and $\sum_{k\in\mathcal K_j}x_{kj}=1$
leaves nothing for $x$ to change, and it leaves costs alone; a row shift instead moves it by
$\sum_jw_jc_{x(j)}$, which depends on the allocation; under \eqref{eq:sat} a column shift moves both sides of the
admissibility test together, while a row shift moves the column best and the admissible set with it.
A shift exceeding the gap between a workload's top two models reverses its ranking, so the
allocation does change. Only the \emph{within-workload spread} of an error is therefore decision
relevant, which lets \eqref{eq:mset} fix each column's level without loss of generality.

\paragraph{The cost of choosing wrongly.} \looseness=-1 For an allocation, written $\pi$ for the
map that sends each workload $j$ to the model $\pi(j)$ chosen for it, the
\emph{target-value loss} is the target quality given up relative to the allocation an operator
holding the true table would make,
\begin{equation}\label{eq:loss}
  \Rref(\pi)=\sum_j w_j\bigl(\muref_{\piref(j),j}-\muref_{\pi(j),j}\bigr),
  \qquad\piref=\Psi[\muref].
\end{equation}
Serving $\pi$ costs $N\sum_j w_jp_{\pi(j),j}$ per period, to which purchased evaluation is added.

\looseness=-1 That loss is written in $\muref$, which the operator does not have. What it has
instead is a \emph{set} $\mathcal U$ of tables the evidence leaves open. This need not be fatal,
since $\Psi$ is piecewise constant and the operator needs the region rather than the point, so the
question is one of containment, written with the decision region
$\Sigma_x=\{\mu:\Psi(\mu)=x\}$ of the tables that produce $x$,
\begin{equation}\label{eq:contain}
  \text{the decision is determined}\qquad\Longleftrightarrow\qquad
  \exists\,x\;:\;\mathcal U\subseteq\Sigma_x
\end{equation}
that is, whether some single allocation $x$ has a region $\Sigma_x$ large enough to hold every table
in $\mathcal U$. If one does, the operator can act without knowing $\muref$, however wide
$\mathcal U$ is. If instead $\mathcal U$ straddles the boundary between two regions the answer is
unknown however narrow it is, and the cells that boundary is written on are the ones worth paying to
measure. \S\ref{sec:model} builds $\mathcal U$,
decides the containment, and says what to buy when it fails.
\section{Proposed approach}\label{sec:model}\label{sec:ident}\label{sec:protocol}

\looseness=-1 The operator's two sources of evidence are the production log and randomized
re-evaluation. We use them to construct $\mathcal U$ rather than to estimate the quality table
exactly.

\subsection{Bounding the quality table}\label{sec:blind}\label{sec:assume}
\looseness=-1 We summarize the production log by the table $M$. Its entry $M_{kj}$ is the average recorded score among
workload-$j$ requests that production assigned to model $k$. We summarize randomized re-evaluation by the
table $R$. Its entry $R_{kj}$ is the average recorded score when workload-$j$ requests are drawn at
random and run on model $k$. We call one purchased randomized observation a \emph{readout} of cell
$(k,j)$.

\looseness=-1 Randomized re-evaluation provides the causal comparison the production log cannot. A model may look
better in the log simply because it was assigned easier requests. Evaluating it on a random sample
of the workload removes that source of bias. The difference between the logged average $M_{kj}$ and
the randomized average $R_{kj}$ measures this selection bias when evaluator error is the same on
average in the two settings. App.~\ref{app:proof} states this as Assumption~\ref{as:recover} and
discusses its limits. Because readouts cover only some cells, we form one correction per model, the average of the
observed differences over that model's readout cells,
$b_k=\operatorname{mean}_{j:\,(k,j)\ \text{read}}\bigl(M_{kj}-R_{kj}\bigr)$. Subtracting it from
every logged entry gives the \emph{selection-adjusted} table $V$, with entries $V_{kj}=M_{kj}-b_k$
and $R_{kj}$ used directly wherever a readout is available.

\looseness=-1 We assume the measurement error left by the correction lies, in workload $j$, between
zero and a width $\eta_j$ that we call the \emph{evaluator band}; Assumption~\ref{as:modellevel} in
App.~\ref{app:proof} is why it is assumed rather than estimated.

\looseness=-1 The evidence therefore leaves the set
\begin{equation}\label{eq:mset}
  \mathcal U=\bigl\{\mu\;\bigm|\;|\mu_{kj}-\hat\mu_{kj}|\le e_{kj}\ \text{ for every }k\in\mathcal K_j\bigr\},
  \qquad \hat\mu_{kj}=R_{kj}-\tfrac{\eta_j}{2},
  \qquad e_{kj}=\tfrac{\eta_j}{2}.
\end{equation}
In words, the randomized score can overstate the designated outcome
by at most one evaluator band, so each cell's true value lies between $R_{kj}-\eta_j$ and $R_{kj}$.
Only the width of that interval matters: by Lemma~\ref{lem:shift} the common $\eta_j/2$ in
\eqref{eq:mset} changes no decision, so placing the band below $R_{kj}$ rather than around it fixes a
level and assumes nothing about the sign of the error. The uncertainty set \eqref{eq:mset} writes
this same interval by its centre and half-width, and the algorithm can therefore work at the level of
$R$ and solve at $V$, which coincides with $R$ when the per-model correction is exact. These are the
quality tables the evidence still leaves open; App.~\ref{app:proof} gives the construction of
$\mathcal U$ in full.

\subsection{Testing whether the evidence determines the allocation}
\looseness=-1 Let $\hat x$ be budget-feasible and optimal at the current estimate $\hat\mu$. For any
quality table $\mu$, $X^\star[\mu]$ is the set of budget-feasible allocations that achieve the
highest quality under that table. We call $\hat x$ \emph{settled} if
\[
\hat x\in X^\star[\mu]\qquad\text{for every }\mu\in\mathcal U.
\]
Among all quality tables still plausible given the production log, the randomized readouts, and
the assumed evaluator band, none makes another affordable allocation better than $\hat x$. Equation~\eqref{eq:contain} called the decision determined only
when every table in $\mathcal U$ leads to the same allocation after applying the fixed tie-breaking
rule. Settlement is slightly weaker. If $\hat x$ ties with another allocation at some table in
$\mathcal U$, it still counts as settled because it remains optimal.

\looseness=-1 Checking every table in $\mathcal U$ appears to require infinitely many solves. For the
budgeted rule, one adverse table is enough. To make $\hat x$ as weak as possible, lower every cell it
selects to the bottom of its interval and raise every other cell to the top. The result is the \emph{adverse
table} $\check\mu$, the table in $\mathcal U$ least favourable to $\hat x$, and this same endpoint
choice is worst against every rival at once,
\begin{equation}\label{eq:adverse}
  \check\mu_{kj}=
  \begin{cases}
    \hat\mu_{kj}-e_{kj}, & k=\hat x(j),\\
    \hat\mu_{kj}+e_{kj}, & \text{otherwise,}
  \end{cases}
\end{equation}
which gives $\hat x$ its smallest possible margin over every competitor.

\begin{proposition}[Exactly when the budgeted allocation is settled]\label{prop:robust}
Let $\hat x$ be budget-feasible and optimal at $\hat\mu$. Then $\hat x$ is settled over
$\mathcal U$ if and only if every budget-feasible $x$ satisfies
\begin{equation}\label{eq:cert}
  \underbrace{\sum_{j:\,x(j)\neq\hat x(j)} w_j\bigl(\hat\mu_{\hat x(j),j}-\hat\mu_{x(j),j}\bigr)}_{\text{margin of }\hat x\text{ over }x}
  \;\ge\;
  \underbrace{\sum_{j:\,x(j)\neq\hat x(j)} w_j\bigl(e_{\hat x(j),j}+e_{x(j),j}\bigr)}_{\text{uncertainty where they differ}},
\end{equation}
and equivalently if and only if $\hat x$ remains optimal at the adverse table $\check\mu$
defined in \eqref{eq:adverse}, that is, $\hat x\in X^\star[\check\mu]$.
\end{proposition}

\noindent\looseness=-1 The inequality compares how much better $\hat x$ is than a competing
allocation $x$ with how much the remaining uncertainty could change that comparison. It is
non-strict because a tie leaves $\hat x$ optimal. Only workloads
where $\hat x$ and $x$ choose different models matter, because workloads where they make the same
choice contribute equally to both allocations. Prop.~\ref{prop:robust} turns this check into two
solves of the same budgeted allocation problem, Eq.~\eqref{eq:mckp}. First find the highest-quality allocation under the
serving budget at $\hat\mu$, giving $\hat x$. Then form the adverse table $\check\mu$, which lowers
the cells selected by $\hat x$ and raises the alternatives as far as $\mathcal U$ allows, and solve
again. If $\hat x$ remains optimal, no table the evidence admits makes another affordable
allocation better, and measurement can stop.

\looseness=-1 If $\hat x$ is no longer optimal, let $\check x$ be the allocation that replaces it.
The cells where they differ form the \textbf{disagreement set}
\begin{equation}\label{eq:boundary}
  \mathcal D=\bigl\{(\hat x(j),j)\ \text{and}\ (\check x(j),j)\ :\ j\ \text{with}\ \check x(j)\neq\hat x(j)\bigr\},
\end{equation}
and only uncertainty in these cells affects the comparison between $\hat x$ and $\check x$.
Evidence bought outside $\mathcal D$ cannot resolve the disagreement. Once it is resolved, the
test runs again because a different allocation may then prevent settlement.

\subsection{Measuring where the allocations disagree}\label{sec:buy}\label{sec:case}

\looseness=-1 A failed test does more than say that the decision is unsettled. It identifies where
additional evidence can change the allocation. CASE next looks for the closest competing allocation
at the current selection-adjusted table $V$. For an alternative model $k$ on workload $j$, let
$v(k,j)$ be the best total quality achievable under the serving budget when $x(j)=k$ is forced, and
let $v(\hat x)$ be the value of the current allocation. Let $j^\star$ be the workload and $k^\star$
the alternative model with the smallest loss from such a forced substitution:
\[
(k^\star,j^\star)\in\arg\min_{j,\ k\in\mathcal K_j\setminus\{\hat x(j)\}}\bigl[v(\hat x)-v(k,j)\bigr].
\]
A smaller gap means that switching workload $j$ to model $k$, while re-optimizing the remaining
workloads, gives up less total quality. Thus, $(k^\star,j^\star)$ identifies the closest competitor
to $\hat x$. The next purchase is an unread cell among $(k^\star,j^\star)$ and
$(\hat x(j^\star),j^\star)$. A purchase does not narrow that cell, whose half-width stays
$\eta_j/2$. It replaces the cell's logged entry with its randomized one and updates the model's
correction $b_k$, moving the centre of $\mathcal U$ and with it the margins the test compares.

\looseness=-1 CASE repeats the loop, updating the quality estimates, running the adverse test, and
buying evidence where it can still change the allocation. Alg.~\ref{alg:audit} gives the full
procedure. CASE stops once the remaining uncertainty cannot change the allocation, without
reading every cell.

\begin{algorithm}[t]
\caption{\textbf{CASE}, with randomized readouts as the evidence bought. Symbols follow
\S\ref{sec:setup} and \S\ref{sec:assume}, $\mathcal L$ the production log, $\mathcal A$ the cells
bought, $n_{\mathrm{read}}$ readouts per cell.}
\label{alg:audit}
\begin{algorithmic}[1]
\Require production log $\mathcal L$, prices $p$, traffic $w$, serving budget $\Gamma$, half-widths $e$, evidence budget $\beta$
\Ensure audit queue, an allocation, its expected quality, its deployment cost, and a settled/unsettled verdict, with the disputed cells $\mathcal D$ on failure
\State $M \gets$ mean logged score per $(k,j)$ in $\mathcal L$,\quad $\mathcal{A} \gets \emptyset$
\For{each model $k$}\Comment{one seed comparison per model}
  \State buy $n_{\mathrm{read}}$ readouts of some $(k,j)$, add the cell to $\mathcal{A}$ and record $R_{kj}$
\EndFor
\While{$|\mathcal{A}| < \beta$ and $\hat x$ is not yet settled}
  \State $b_k \gets \operatorname{mean}_{(k,j)\in\mathcal{A}} (M_{kj} - R_{kj})$ for each $k$
  \State $V_{kj} \gets M_{kj} - b_k$, with $R_{kj}$ substituted where bought
  \State $\hat x \gets X^\star[V]$ by solving \eqref{eq:mckp} at $V$ under budget $\Gamma$
  \State $\check V_{kj} \gets V_{kj} - e_{kj}$ if $\hat x$ selects $(k,j)$, else $V_{kj} + e_{kj}$
        \Comment{the adverse table}
  \State $\check x \gets X^\star[\check V]$
  \If{$\hat x \in X^\star[\check V]$} \textbf{return} $\hat x$, \emph{settled}
        \Comment{Prop.~\ref{prop:robust}}
  \EndIf
  \State $\mathcal D \gets \{(\hat x(j),j),(\check x(j),j) : \check x(j)\neq\hat x(j)\}$
        \Comment{the cells in dispute, \eqref{eq:boundary}, reported with the verdict}
  \State buy $n_{\mathrm{read}}$ readouts of one unread cell by the closest-competitor rule of \S\ref{sec:buy}
\EndWhile
\State \Return $\mathcal{A}$, $\hat x$, $\sum_j w_j V_{\hat x(j),j}$, $N\sum_j w_j p_{\hat x(j),j}$,
        and \emph{unsettled} with the disputed cells $\mathcal D$: the evidence budget was exhausted
        before the test of Prop.~\ref{prop:robust} passed
\end{algorithmic}
\end{algorithm}


\section{Experiments}\label{sec:results}
\looseness=-2 We ask whether the selection correction of \S\ref{sec:assume} recovers
decision-relevant quality, whether the certificate of Prop.~\ref{prop:robust} settles the allocation
at the evaluator widths real data exhibits, and whether closing the remaining gap is worth paying
for. It does, it seldom does, and it is. On the one production log carrying both a proxy score and
the outcome it stands in for, CASE gives up $1.57$ of target value against $1.76$ for trusting the
log and $1.94$ for a traffic vote, all in units of $10^{-3}$ (Table~\ref{tab:kuairand}). On LLMRouterBench, under faithful scoring, it removes $32\%$ of the loss that trusting the
log incurs, which is $89\%$ of what the same correction reaches with unlimited audit information,
and it needs none of the logging propensities that no corpus we study records. The certificate is exact
and costs one extra solve, yet at the band we measure on real data it settles $1$ of $210$ instances:
the binding constraint is the evidence, not the optimizer, and better evidence is worth $0.046$ of
target quality per request on LLMRouterBench.

\looseness=-1 Every corpus but KuaiRand is an offline replay of a benchmark grid that records each
model's score on each prompt, so the quality table is known and the target-value loss \eqref{eq:loss}
of any allocation can be computed. A workload is a group of prompts, and a \emph{partition} is one
random draw of the disjoint cohorts a run needs: one to fit the router that produces the production
log, one to serve as that log, one to draw readouts from, and one to score the allocation each method
returns. Every number below averages over $30$ such draws and carries one standard error. Two budgets
appear throughout. The serving budget $\Gamma$ of \eqref{eq:mckp} is reported as a multiple of the
cheapest feasible spend, and the \emph{evidence budget} is the share of model--workload cells the
audit may buy, $35\%$ unless stated otherwise. KuaiRand is a production log rather than a replay,
and \S\ref{sec:kuairand} takes it first. App.~\ref{app:extra} gives the design checks, interval
construction, and causal baselines.

\subsection{What CASE recovers on a production log}\label{sec:kuairand}
\looseness=-1 KuaiRand-Pure~\citep{gao2022kuairand} is the one corpus whose requests carry both a
click, which we treat as the recorded proxy score, and a like, which we treat as the designated
outcome, so a deployed decision can be priced in the outcome the operator values and errors caused
by selective routing separate from errors caused by the score itself. Table~\ref{tab:kuairand}
reports the target-value loss of four ways of deciding, among them a traffic vote that never
consults the evaluator.

\begin{table}[t]\centering\scriptsize\setlength{\tabcolsep}{5pt}
\caption{\textbf{Removing assignment bias does not remove measurement loss.} Target-value loss is
measured in the KuaiRand like outcome ($\times10^{3}$) with a $35\%$ evidence budget, reported as
mean $\pm$ one standard error over $30$ log partitions. The traffic vote assigns each workload to
its most-used model and does not use the evaluator. Under randomized assignment there is no
most-used model to vote for, so those cells are dashed. Allocations use the quality-constrained
rule~\eqref{eq:sat} with $\delta=0$. Target-outcome scoring is unbiased by construction but noisier
because likes are $36$ times rarer than clicks.}
\label{tab:kuairand}
\begin{tabular}{@{}lcccc@{}}
\toprule
& \multicolumn{2}{c}{evaluator = click (D1)} & \multicolumn{2}{c}{evaluator = target (D0)}\\
\cmidrule(lr){2-3}\cmidrule(l){4-5}
Deployed decision & A1 (selective) & A0 (randomized) & A1 (selective) & A0 (randomized)\\
\midrule
Trust-the-log        & $1.76\pm0.14$ & $1.49\pm0.07$ & $1.76\pm0.10$ & $1.94\pm0.08$\\
CASE                 & $1.57\pm0.06$ & $1.60\pm0.06$ & $1.74\pm0.06$ & $1.72\pm0.06$\\
Traffic vote         & $1.94\pm0.04$ & ---           & $1.94\pm0.04$ & ---\\
Cellwise-perfect repair & $2.01\pm0.02$ & $2.02\pm0.02$ & $0.00$        & $0.00$\\
\bottomrule
\end{tabular}
\end{table}

\looseness=-2 The fourth row is an oracle. A \emph{cellwise-perfect repair} removes the selection
bias separately in every model--workload cell, showing what randomized re-evaluation could achieve
if that bias were corrected exactly, so what it leaves is due to the mismatch between clicks and
likes alone. It leaves $2.01$, more than CASE: the router's preference for some models and the click
score's preference partly cancel, so removing only the first exposes the second.
App.~\ref{app:extra} gives the full comparison and checks this interaction.

\looseness=-2 The same corpus measures the band that \S\ref{sec:rq3} tests against. Comparing the
click score with the like outcome in each cell, the spread across models within a workload is the
$\eta_j$ of \S\ref{sec:assume}: median $0.128$ over the eight workloads, ranging from $0.095$ to
$0.173$, more than six times the $0.02$ used in our benchmark replays. Real judges are no narrower,
at $0.336$ on LMArena (App.~\ref{app:extra}, Table~\ref{tab:meas}).

\subsection{Randomized re-evaluation repairs selection but leaves a floor}\label{sec:rq2}
\looseness=-1 Evaluator error survives the removal of selection bias, so we ask whether randomized
re-evaluation still helps choose the allocation. We vary whether requests are selectively or
randomly assigned and whether the evaluator is faithful or distorted, across serving budgets.

\begin{figure}[htb]\centering
\includegraphics[width=0.88\textwidth]{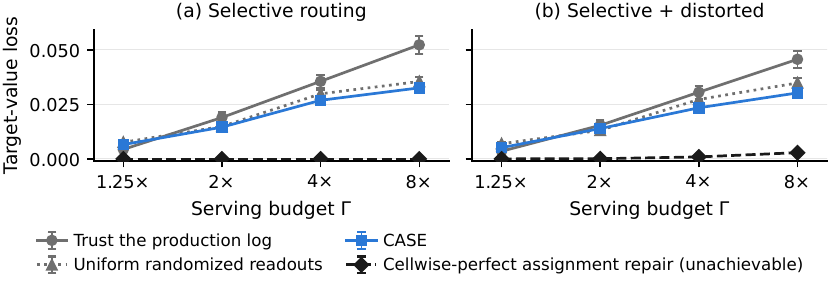}
\caption{\textbf{Randomized evidence repairs selection but leaves the evaluator floor.} Target-value
loss of \eqref{eq:mckp} against $\Gamma$ on LLMRouterBench~\citep{li2026llmrouterbench}, with a $35\%$ evidence budget and
$\pm$ SE over $30$ partitions $\times$ $6$ seeds. Dashed is the unachievable \emph{Cellwise-perfect
assignment repair}.}
\label{fig:validity}
\end{figure}

\looseness=-2 Figure~\ref{fig:validity} shows how much target quality each method gives up at each
serving budget, first under selective routing alone and then with a distorted evaluator added. With
selective routing, CASE reduces target-value loss at $2\times$, $4\times$ and $8\times$ the cheapest
feasible spend; at $1.25\times$ the budget itself fixes most of the allocation and the correction does
not pay for itself. With randomized assignment there is no selection bias to remove, and CASE no
longer improves on trusting the log. A distorted evaluator leaves additional loss
under either assignment scheme, consistent with the residual uncertainty captured by $\eta_j$. RouterBench~\citep{hu2024routerbench} shows the same selection
correction at three of the four budgets tested. The correction is also near its own ceiling:
against the raw routed log's $0.0499$, CASE leaves $0.0338$ and the best correction of the same form
with unlimited audit information leaves $0.0318$, so under faithful scoring a $35\%$ evidence budget
already buys $89\%$ of what more evidence of this kind could buy (App.~\ref{app:extra}).

\looseness=-1 Selective routing can also be corrected from the production log when assignment
probabilities are recorded. Using the true probabilities, log-only estimators match CASE without
audit spending. None of the corpora we study records those probabilities, so this correction is not
available here, and estimating them instead performs worse. App.~\ref{app:causal} gives the estimators and full results.

\subsection{When the range determines the allocation}\label{sec:rq3}
\looseness=-1 We measure how much evaluator uncertainty each instance tolerates before its optimal
allocation changes. For each partition and serving budget we increase a common evaluator spread
$\eta$ until the current allocation is no longer optimal throughout $\mathcal U$, and call the
largest tolerable value the \emph{decision margin}.

\looseness=-2 Table~\ref{tab:cert} reports the decision margins by serving budget. On
LLMRouterBench, the median margin is $0.0456$ at $1.25\times$ the cheapest feasible spend, with $21$ of $30$ partitions tolerating $\eta=0.02$. At
$8\times$, the median falls to $0.0026$ and none do. Tight budgets rule out many expensive
alternatives, leaving fewer ways for uncertainty to change the allocation. RouterBench shows a
weaker pattern, and neither benchmark is monotone across all seven budgets. To check whether this
result depends on allowing evaluator error to vary across model--workload cells, we repeat the
analysis with one common error per model across workloads. The decision margins increase by about
one third, but only one of $420$ verdicts changes at the measured width. App.~\ref{app:extra} gives
the full budget sweep and this structural check.

\begin{table}[t]\centering\scriptsize\setlength{\tabcolsep}{5pt}
\caption{\textbf{The range is more likely to determine the allocation under a tight serving budget on
LLMRouterBench.} The decision margin is the largest evaluator spread $\eta$ for which the allocation
stays optimal over all of $\mathcal U$, with counts of the $30$ partitions reaching two spreads.}
\label{tab:cert}
\begin{tabular}{@{}llccc@{}}
\toprule
Corpus & $\Gamma$ / cheapest & median margin & margin $\ge0.01$ & margin $\ge0.02$\\
\midrule
LLMRouterBench & $1.25\times$ & $0.0456$ & $23/30$ & $21/30$\\
LLMRouterBench & $2\times$    & $0.0179$ & $20/30$ & $14/30$\\
LLMRouterBench & $8\times$    & $0.0026$ & $5/30$  & $0/30$\\
\addlinespace[2pt]
RouterBench    & $1.25\times$ & $0.0132$ & $16/30$ & $10/30$\\
RouterBench    & $2\times$    & $0.0133$ & $18/30$ & $9/30$\\
RouterBench    & $8\times$    & $0.0037$ & $6/30$  & $1/30$\\
\bottomrule
\end{tabular}
\end{table}

\looseness=-2 These margins are small relative to the evaluator error measured in real data.
KuaiRand has median $\eta=0.128$, nearly three times the largest median margin in
Table~\ref{tab:cert}, and only one of the $210$ LLMRouterBench partition--budget pairs remains
determined at that width. Thus randomized re-evaluation can improve the estimate of $\muref$
without fixing the allocation. CASE therefore buys targeted evidence for the model--workload cells
that can still change the decision, as described in \S\ref{sec:buy}.

\subsection{What better information about the quality table buys}\label{sec:rq5}
\looseness=-1 We now price that uncertainty. Holding the serving budget at $\Gamma=8\times$, we solve
\eqref{eq:mckp} from the raw production log, from the selection-adjusted table, and from the
designated outcome, then score all three allocations on the designated outcome. Target quality rises
$0.4798\to0.4924\to0.5255$ and deployment spend rises with it, \$$3{,}109\to$\,\$$3{,}850\to$\,\$$4{,}386$
per million requests against a \$$4{,}557$ cap: the budget is a cap rather than a target, so an
inaccurate table underspends it, sending work to cheap models that a correct table would not have
chosen. The selection-adjusted table recovers about one quarter of the quality gap, and assigns a
different model from the designated-outcome table on $59.3\%$ of workloads against $63.7\%$ for the
raw log. App.~\ref{app:measured} tabulates all three.

\looseness=-1 The value is also substantial in dollars. On $150$ paid
SWE-Lancer~\citep{miserendino2025swelancer} tasks, routing with the observable quality table gains
about \$1{,}000 over the best single model, while an oracle routing on each task's realized outcome
gains a further \$35{,}000. That oracle also decides per task rather than per payout band, so the gap
bounds the value of better quality information rather than measuring it.
App.~\ref{app:measured} reports both comparisons in full. The allocation is
solved exactly at every table, so a better optimizer applied to the same table has nothing left to
gain, and the value lies in better measurement.


\section{Conclusion}\label{sec:conclusion}
\looseness=-2 An operator must assign each workload to one model under a spending cap. If the average
target quality of every model on every workload were known, collectively forming the quality table
$\muref$, the assignment would be a standard knapsack problem. The difficulty is that this table is
unknown. Randomized re-evaluation can correct selective assignment, but it does not touch the
evaluator's mean departure from the designated outcome. We therefore bound the quality tables consistent with the evidence and ask whether they all
imply the same allocation. For the budgeted rule, two solves are enough. Agreement settles the
decision, while disagreement identifies the model--workload pairs where more evidence can matter. At
the evaluator bands we measure, the method often declines to certify, showing that trustworthy
quality information, not optimization, is the binding constraint.

\paragraph{Limitations.} \looseness=-2
CASE certifies robustness only to the uncertainty set $\mathcal U$ supplied to it, not that the
unknown table $\muref$ lies in that set, and the sampling error of the estimated centre of
$\mathcal U$ sits outside it. Our row-level correction can also miss cell-specific residual bias,
especially for cells that are never re-evaluated. Most evaluations are offline replays rather than
live deployments. And where designated outcomes are unavailable the band must be assumed, and
\S\ref{sec:kuairand} shows the error in real data can be far larger.

\smallskip\noindent\textbf{Reproducibility.}
\url{github.com/HamedKhosravi99/logged-llm-comparison-audit}.

\label{endofmaintext}
\bibliographystyle{plainnat}
{\footnotesize
\setlength{\bibsep}{2.5pt}
\bibliography{tae_refs}}

\appendix
\newpage
\small

\section{Notation}\label{app:notation}
Table~\ref{tab:notation} lists the symbols the main text uses, in the order
\S\ref{sec:setup} and \S\ref{sec:model} introduce them.

\begin{table}[H]\centering\scriptsize\setlength{\tabcolsep}{5pt}
\caption{Notation, in the order this section introduces it. The tables $\muref$, $M$ and $R$ differ only in
\emph{who was measured} and \emph{what the measurement was}. Only $M$ and $R$ are observable, and
only through estimates. The evaluator gap $C$ and its parts are introduced in App.~\ref{app:proof}
and used only in the appendices.}
\label{tab:notation}
\begin{tabular}{@{}llp{7.9cm}@{}}
\toprule
Symbol & Name & What it is\\
\midrule
\multicolumn{3}{@{}l@{}}{\emph{Primitives} (\S\ref{sec:dgp})}\\
$w_j$, $p_{kj}$ & share, price & workload $j$'s traffic share, and model $k$'s per-request price on it\\
$N$, $\Gamma$ & volume, serving budget & requests per budget period, and the spend allowed on them\\
$\mathcal K_j$ & candidate set & models supported on workload $j$\\
\midrule
\multicolumn{3}{@{}l@{}}{\emph{Tables} (\S\ref{sec:obs}, \S\ref{sec:assume})}\\
$\muref_{kj}$ & target-outcome table & average designated outcome of model $k$ over \emph{all} workload-$j$ requests; the unknown\\
$M_{kj}$ & logged score table & expected recorded score of model $k$ over the workload-$j$ requests \emph{the router sends it}\\
$R_{kj}$ & randomized-audit table & expected recorded score of model $k$ on a fresh \emph{random} sample of workload $j$\\
$\hat M_{kj}$, $\hat R_{kj}$ & estimates & sample means over $n^{\mathrm{log}}_{kj}$ logged and $n_{kj}$ randomized scores\\
$b_k$, $V_{kj}$ & correction, selection-adjusted table & one offset per model from the bought cells, and $M_{kj}-b_k$ with readouts substituted where bought\\
\midrule
\multicolumn{3}{@{}l@{}}{\emph{Error and uncertainty} (\S\ref{sec:obs}, \S\ref{sec:assume}, App.~\ref{app:proof})}\\
$\Bsel_{kj}$ & selection gap & how much easier or harder the router's requests were, in target units\\
$C_{kj}$ & evaluator gap & how far the recorded score sits from the designated outcome, per cell; App.~\ref{app:proof} only\\
$C_k$ & model-level gap & the case $C_{kj}=C_k$, the evaluator's mean error not varying across workloads; appendices only\\
$\eta_j$ & evaluator band & assumed width of the interval \eqref{eq:mset} that $\muref_{kj}$ lies in, below $R_{kj}$\\
$\mathcal P$ & model-level class & the processes satisfying Assumptions~\ref{as:recover} and~\ref{as:modellevel}\\
\midrule
\multicolumn{3}{@{}l@{}}{\emph{Decision and criterion} (\S\ref{sec:rules}, \S\ref{sec:ident})}\\
$X^\star[\mu]$, $\pidel[\mu]$ & the two rules & budgeted allocation \eqref{eq:mckp}, and quality-constrained rule \eqref{eq:sat} at tolerance $\delta$\\
$\Psi[\mu]$ & either rule & written where a statement holds for both\\
$\piref$ & target decision & $\Psi[\muref]$, the decision the true table would produce\\
$\Sigma_x$, $\tau(\mu)$ & decision region, stability radius & the tables that produce $x$, and the largest perturbation of $\mu$ that leaves $\Psi$ unchanged\\
$\mathcal U$ & the box & the tables the evidence admits, \eqref{eq:mset}; the decision is \emph{settled} when one allocation is optimal at every one of them\\
$e_{kj}$ & half-width & radius of cell $(k,j)$ in $\mathcal U$, half the evaluator band\\
$\mathcal D$ & disagreement set & the cells where the incumbent and its challenger differ, \eqref{eq:boundary}\\
$\Rref(\pi)$ & target-value loss & target value given up by deploying $\pi$ instead of $\piref$\\
$\beta$ & evidence budget & model--workload cells the operator may buy, at $n_{\mathrm{read}}$ readouts each\\
$D_{kj}$ & observable gap & distance from model $k$ to the best other model in workload $j$, on $R$, in the appendices only\\
\bottomrule
\end{tabular}
\end{table}

\section{Proofs}\label{app:proof}
This appendix collects the formal content deferred from \S\ref{sec:setup} and \S\ref{sec:model}, namely the
request-level setup the statements below share, the
counterexamples separating the two decision rules, the proof of the
two conditions on the evaluator that the construction assumes, with Cor.~\ref{cor:blind},
and the proof of Prop.~\ref{prop:robust}.

\paragraph{Setup.} A request carries a workload label $Z$ and was served by a model
$A\in\mathcal K_Z$, and $\Yref(k)$ is the designated outcome it would receive if model $k$ served it,
defined for every model and not only the one that did \citep{imbens2015causal}, so that
$\muref_{kj}=\mathbb E[\Yref(k)\mid Z=j]$. Write $Y^o(k)$ and $Y^r(k)$ for the recorded scores under
the observational and the randomized design,
$C^{o}_{kj}=\mathbb E[Y^o(k)-\Yref(k)\mid A{=}k,Z{=}j]$ and
$C^{r}_{kj}=\mathbb E[Y^r(k)-\Yref(k)\mid Z{=}j]$ for the evaluator gap under each, and
$\Bsel_{kj}=\mathbb E[\Yref(k)\mid A{=}k,Z{=}j]-\muref_{kj}$ for the selection gap, which vanishes
when $\Yref(k)\perp A\mid Z=j$. The two observables are then
\begin{equation}\label{eq:twosurfaces}
  M_{kj}=\muref_{kj}+\Bsel_{kj}+C^{o}_{kj},
  \qquad
  R_{kj}=\muref_{kj}+C^{r}_{kj},
\end{equation}
an identity in these definitions rather than an assumption, and $\mathcal U$ of \eqref{eq:mset} is
the set of tables $R-C$ over the error tables
\begin{equation}\label{eq:cset}
  \mathcal C=\bigl\{\,C\;\bigm|\;C_{kj}\in[0,\eta_j]\ \text{for every }k\in\mathcal K_j\text{ and every }j\,\bigr\}.
\end{equation}
Neither rule reads the level of a column, so only the within-workload spread of $C$ is decision
relevant and \eqref{eq:cset} loses nothing by placing that spread in $[0,\eta_j]$.

\noindent Two conditions on the evaluator carry the construction. We assume them rather than
establish them.

\begin{assumption}[Assignment recovery]\label{as:recover}
The evaluator's mean error is the same in the two streams, $C^{o}_{kj}=C^{r}_{kj}=:C_{kj}$ in every
cell.
\end{assumption}

\begin{assumption}[Model-level evaluator error]\label{as:modellevel}
That error does not vary across workloads, $C_{kj}=C_k$. Write $\mathcal P$ for the processes
satisfying both conditions.
\end{assumption}

\noindent Subtracting the two lines of \eqref{eq:twosurfaces} under Assumption~\ref{as:recover}
leaves $M_{kj}-R_{kj}=\Bsel_{kj}$ in every cell, which is the correction \S\ref{sec:assume} applies;
what it then targets is $R_{kj}$ rather than $\muref_{kj}$. The condition constrains evaluator error
and not assignment, and App.~\ref{app:tests} measures its residual. Using one evaluator on both
streams does not deliver it: that equates the two error \emph{functions}, not the two conditional
means, which are taken over different populations whenever the router selects within a workload
(Rem.~\ref{rem:samejudge}).

\noindent Under Assumption~\ref{as:modellevel} the evaluator's error cancels from $M-R$ identically
(Cor.~\ref{cor:blind}), so the per-model offset $b_k$ of \S\ref{sec:assume} estimates assignment bias
and cannot remove $C_k$, even though both have the same functional form. Moving a constant $c_k$ out
of the designated outcome and into the evaluator error leaves the recorded score, the designated
outcome plus the evaluator's error on that request, the same random variable, so the production log
and a randomized evaluation of any size have the same joint law before and after the move. Within
$\mathcal P$ a model-level evaluator preference is therefore indistinguishable from an equal
difference in target quality, $\muref$ is
pinned down only up to a per-model shift, and $\eta_j$ is assumed in \eqref{eq:mset} rather than
estimated.

\paragraph{The two decision rules do not reduce to one another.} Both instances below are exact in
rationals. First, rule~\eqref{eq:sat} can select a model that no linear scalarization of
\eqref{eq:mckp} reaches, already with a single workload. Take three models of quality
$(1.00,\,0.99,\,0.50)$ priced $(10,\,9.9,\,1)$ with $\delta=0.02$. The rule admits the first two and
buys the second. That model maximizes $\mu_k-\lambda p_k$ for no $\lambda\ge0$, since beating the first
model requires $\lambda>1/10$ and beating the third requires $\lambda<49/890$. It is
Pareto-efficient but interior to the convex hull of the price--quality frontier, and linear
scalarization reaches only that hull, which is why the greedy and Lagrangian routes to
\eqref{eq:mckp} cannot produce it.

Second, the selection of \eqref{eq:sat} need not be budget-optimal at its own spend, because
\eqref{eq:mckp} can trade quality across workloads and \eqref{eq:sat} cannot. Take two equally sized
workloads, the first offering qualities $(1.00,\,0.97,\,0.96)$ at prices $(10,2,1)$ and the second
$(0.99,\,0.97)$ at prices $(3,2)$, with $\delta=0.03$. Rule~\eqref{eq:sat} buys the $0.97$ model in
each, spending $4$ for weighted quality $1.94$. At $\Gamma=4$ program~\eqref{eq:mckp} instead buys
$0.96$ and $0.99$, for $1.95$ --- and that allocation puts the first workload $0.04$ below its best
model, violating the very tolerance \eqref{eq:sat} exists to enforce. So neither is a relaxation of
the other in either direction.

\begin{remark}[Why using the same evaluator is not enough]\label{rem:samejudge}
Using the same evaluator on both streams is insufficient when routing changes the mix of requests on
which that evaluator makes errors, and additional assumptions are then required.
\end{remark}

\begin{corollary}[Blindness to model-level evaluator preference]\label{cor:blind}
If $P\in\mathcal P$, so the evaluator's mean error is model-level, then $M_{kj}-R_{kj}=\Bsel_{kj}$
for every cell, since the distortion cancels identically from the logged-minus-randomized comparison.
\end{corollary}

\paragraph{Proof of Prop.~\ref{prop:robust}.} Write $\mathcal F$ for the set of budget-feasible
allocations. The constraint $N\sum_j w_j p_{x(j),j}\le\Gamma$ involves prices and traffic only, so
$\mathcal F$ does not depend on $\mu$. Put $\mathcal V(x;\mu)=\sum_j w_j\mu_{x(j),j}$ and
$\mathcal J(x)=\{j:x(j)\neq\hat x(j)\}$.
\emph{Agreeing cells cancel.} For every $\mu$ and every $x$,
\[
  \mathcal V(\hat x;\mu)-\mathcal V(x;\mu)=\sum_j w_j\bigl(\mu_{\hat x(j),j}-\mu_{x(j),j}\bigr)
  =\sum_{j\in \mathcal J(x)} w_j\bigl(\mu_{\hat x(j),j}-\mu_{x(j),j}\bigr)=:\Phi(x;\mu),
\]
since each $j\notin \mathcal J(x)$ contributes $\mu_{\hat x(j),j}-\mu_{\hat x(j),j}=0$.
\emph{The per-cell worst case is attainable.} Fix $x$. For $j\in \mathcal J(x)$ the cells $(\hat x(j),j)$ and
$(x(j),j)$ are distinct, and cells for different $j$ lie in different columns, so the $2|\mathcal J(x)|$
entries appearing in $\Phi(x;\cdot)$ are distinct coordinates of $\mu$. That map is affine with
coefficient $+w_j>0$ on $\mu_{\hat x(j),j}$ and $-w_j<0$ on $\mu_{x(j),j}$, and $\mathcal U$ is a
product of intervals, so each coordinate may be set to its own extreme independently and
\[
  \min_{\mu\in\mathcal U}\Phi(x;\mu)=\sum_{j\in \mathcal J(x)} w_j\bigl[(\hat\mu_{\hat x(j),j}-e_{\hat x(j),j})
  -(\hat\mu_{x(j),j}+e_{x(j),j})\bigr].
\]
Now $\hat x$ is settled exactly when $\min_{\mu\in\mathcal U}\Phi(x;\mu)\ge0$ for every $x\in\mathcal F$,
which on rearranging is \eqref{eq:cert}.
\emph{One solve suffices.} Let $\check\mu_{kj}=\hat\mu_{kj}-e_{kj}$ for $k=\hat x(j)$ and
$\hat\mu_{kj}+e_{kj}$ otherwise, so $\check\mu\in\mathcal U$. For any $x$ and any $j\in \mathcal J(x)$ we have
$x(j)\neq\hat x(j)$, so $\check\mu$ lowers $(\hat x(j),j)$ and raises $(x(j),j)$, which is the
minimizing assignment above. Hence $\Phi(x;\check\mu)=\min_{\mu\in\mathcal U}\Phi(x;\mu)$ for every $x$ at
once, and $\hat x$ is settled iff $\Phi(x;\check\mu)\ge0$ throughout $\mathcal F$, iff
$\hat x\in X^\star[\check\mu]$. \qed
Both directions and the equivalence were additionally checked against brute force over $31{,}724$
sampled instances spanning lattice and continuous coefficients, wide error bounds and the degenerate
exact-coefficient case.

\noindent The appendices use a few symbols beyond Table~\ref{tab:notation}, namely the request-level
$Z$, $A$ and $\Yref(k)$ of the setup above, and the measurement floor $\Gmeas$ that a
cellwise-perfect repair leaves behind.

\paragraph{Remark~\ref{rem:samejudge}, continued.} Our stream-mismatch check fails
Assumption~\ref{as:recover}, a mismatch App.~\ref{app:extra} measures at $0.0018$ against a selection
gap of $0.0991$. Our nominal distortion is cell-constant, so the assumption holds of it. Clipping
makes the \emph{realized} error slightly request-dependent, and we measure that residual instead of assuming it away (\S\ref{sec:results}, App.~\ref{app:tests}).

\section{The quality-constrained rule}\label{app:sat}
This appendix carries the analysis of the quality-constrained rule $\pidel$ of \eqref{eq:sat}, which
fixes a per-workload quality standard and minimizes cost instead of fixing spend and maximizing
quality. Because \eqref{eq:sat} decomposes --- workload $j$ is settled by column $j$ of $\mu$ alone
--- its identification question is answered workload by workload rather than by the global
certificate Prop.~\ref{prop:robust} supplies for \eqref{eq:mckp}. The result below is the analogue of
that proposition, and \S\ref{sec:criterion} gives the pair of outcomes it is judged by.

\begin{proposition}[When the quality-constrained decision is point-identified]\label{prop:idset}
Under Assumption~\ref{as:recover} and the admissible error set $\mathcal C$
of \eqref{eq:cset}, fix a workload $j$ with at least two candidate models. Taking
\emph{admissible} as in \S\ref{sec:rules}, call a model \emph{ambiguous} when it is admissible under
some $C\in\mathcal C$ but not all. Then
\begin{equation}\label{eq:trichotomy}
  \text{model }k\text{ is}\quad
  \begin{cases}
    \text{admissible under every }C\in\mathcal C, & D_{kj}\le\delta-\eta_j,\\[1pt]
    \text{admissible under no }C\in\mathcal C,    & D_{kj}>\delta+\eta_j,\\[1pt]
    \text{ambiguous},                             & \text{otherwise.}
  \end{cases}
\end{equation}
The decision $\pidel$ is point-identified in workload $j$ if and only if a universally admissible
model exists and no ambiguous model precedes the cheapest such model in the cost and tie-breaking
order. In particular, $|D_{kj}-\delta|>\eta_j$ for every $k$ is sufficient.
\end{proposition}

\paragraph{Proof of Prop.~\ref{prop:idset}.} Fix $j$, drop it from the notation, and write
$\gamma_k(C)=\max_{k'}(R_{k'}-C_{k'})-(R_k-C_k)$ for the gap of model $k$ on the candidate target
table $R-C$, so $k$ is admissible iff $\gamma_k(C)\le\delta$. Since the $k'=k$ term contributes $0$,
$\gamma_k(C)=\max\bigl(0,\;\max_{k'\neq k}\{(R_{k'}-R_k)-(C_{k'}-C_k)\}\bigr)$, and
$\max_{k'\neq k}(R_{k'}-R_k)=D_k$ by definition. Each difference $C_{k'}-C_k$ ranges over $[-\eta,\eta]$, and the
choices $C_{k'}=C_k\mp\eta$ for all $k'\neq k$ are jointly feasible, so the two extremes are attained
simultaneously and $\gamma_k$ ranges over $[\max(0,D_k-\eta),\,\max(0,D_k+\eta)]$. As $\delta\ge0$, model
$k$ is admissible for every $C$ iff $D_k+\eta\le\delta$ and for no $C$ iff $D_k-\eta>\delta$, which is
the stated classification.

For the decision, order models by price and let $k^\dagger(C)$ be the cheapest admissible one, so point
identification means $k^\dagger$ is constant on $\mathcal C$. If some model is always admissible, let $k_0$
be the cheapest such. Then $k^\dagger(C)\le k_0$ in price for every $C$. If no ambiguous model is cheaper
than $k_0$, every model cheaper than $k_0$ is admissible under no $C$, so $k^\dagger\equiv k_0$. Conversely,
suppose some ambiguous model is cheaper than $k_0$ and let $k_1$ be the cheapest such. Ambiguity
gives $C$ with $k_1$ admissible, whence $k^\dagger(C)\le k_1$, and $C'$ with $k_1$ inadmissible, whence
$k^\dagger(C')\neq k_1$. Were $k^\dagger$ constant at $k_2$, both would force $k_2$ strictly cheaper than $k_1$ and
$k_2$ admissible under $C'$. But every model cheaper than $k_1$ is, by the choice of $k_0$ and $k_1$,
admissible under no $C$, a contradiction. Finally, if no model is always admissible, take any $C$ and
put $k_2=k^\dagger(C)$, so $k_2$ is admissible somewhere, hence ambiguous, and some $C'$ makes it
inadmissible with $k^\dagger(C')\neq k_2$. Both directions were also checked by brute force against the achievable winner sets
over a lattice of instances chosen to realize the boundary cases and price ties.


\section{Experimental details and additional results}\label{app:tests}\label{app:extra}\label{app:ci}
\label{app:econ}\label{app:causal}\label{app:measured}\label{app:kuairand}\label{app:report}
\label{app:cert}\label{app:costred}

The appendix results follow the same empirical question as \S\ref{sec:results}. Randomized evidence should repair the part of the observed quality table created by selective routing, while evaluator error should remain as width around $\muref$. That distinction matters when the resulting set $\mathcal U$ can still cross a decision boundary. The value of further measurement is then the value of resolving those remaining crossings. The results below add the checks and secondary comparisons needed to see where this sequence holds and where it weakens.

\looseness=-1 The decision rule is not the same throughout. The margin sweep below solves the
budgeted allocation of \eqref{eq:mckp}, as do \S\ref{sec:rq2}, \S\ref{sec:rq3} and
\S\ref{sec:rq5}. The other runs here, and the production-log replay of \S\ref{sec:kuairand}, use
a per-workload argmax with no shared budget, which is the quality-constrained rule of
App.~\ref{app:sat} at $\delta=0$. Levels are therefore not comparable cell by cell across the two,
and only directions are offered as robustness.

\subsection{Separating what randomized evidence can and cannot repair}

The controlled replay is informative only if routing and scoring error can move separately. We therefore cross selective or matched-random assignment with faithful or distorted scoring, keep the fit, audit, and scoring cohorts disjoint, and give paired assignment conditions the same per-cell counts. Replacing the designated outcomes available only to the scoring code leaves the acquired cells, fitted correction, and candidate allocation unchanged. The remaining mismatch between evaluator error in the logged and randomized streams is $0.0018\se{0.0001}$, compared with selection gap $0.0991\se{0.0019}$. The replay therefore creates a much larger assignment effect while still measuring the residual mismatch rather than assuming it away.

Once the two effects are separated, the next question is whether CASE is limited by too few randomized readouts or by uncertainty those readouts cannot remove. We compare the deployed one-offset-per-model correction with the best correction of the same form given unlimited audit information and with an unachievable cellwise repair that removes assignment error everywhere. Table~\ref{tab:ceiling} shows that the deployed correction is already close to the best row correction. The cellwise repair reaches zero loss under faithful scoring but leaves $0.0030\se{0.0006}$ under distorted scoring. More randomized evidence can move CASE toward the correction-family ceiling, but it cannot remove the part of the error that comes from what the evaluator measures.

\begin{table}[htbp]\centering\scriptsize\setlength{\tabcolsep}{5pt}
\caption{Correction-family ceilings at a $35\%$ evidence budget over $30$ partitions. The cellwise repair removes assignment error everywhere. It reaches zero target-value loss under faithful scoring and leaves $0.0030$ under distorted scoring.}
\label{tab:ceiling}
\begin{tabular}{@{}llcc@{}}
\toprule
Correction surface & Deployable? & A1--D0 & A1--D1\\
\midrule
Raw routed log & yes & $0.0499\se{0.0030}$ & $0.0469\se{0.0025}$\\
CASE row offset & yes & $0.0338\se{0.0020}$ & $0.0338\se{0.0017}$\\
Oracle row offset, no budget limit & no & $0.0318\se{0.0025}$ & $0.0337\se{0.0024}$\\
Cellwise-perfect assignment repair & no & $0.0000$ & $0.0030\se{0.0006}$\\
\bottomrule
\end{tabular}
\end{table}

If that split is a property of the mechanism rather than one benchmark, it should survive changes in the request distribution, candidate set, and distortion. On RouterBench, with $11$ candidate models, CASE again repairs part of the selective-routing loss. Under faithful scoring the loss falls from $0.0784$ for the routed log to $0.0421$, while under distorted scoring the cellwise-perfect assignment repair still leaves $0.0053$. Doubling the distortion raises the wrong-model floor from $18.3\%$ to $30.6\%$ and the remaining target-value loss from $0.0030$ to $0.0069\se{0.0014}$. A model preference constant across workloads is even harder for differencing to see, yet the cellwise-perfect repair still selects the wrong model on $10.0\%\se{2.1}$ of workloads. Two secondary effects do not survive. The acquisition-rule ordering reverses with the larger candidate set and the cost of auditing under matched-random assignment changes sign. We therefore leave those effects out of the main claim.

Randomized re-evaluation is also not the only way to repair selective routing. If assignment propensities are recorded, the selective log can be reweighted instead. We run outcome regression, stabilized H\'ajek inverse propensity weighting \citep{swaminathan2015self}, and cross-fitted augmented inverse propensity weighting \citep{dudik2011doubly} on the same partitions. Table~\ref{tab:causalbase} shows that exact-propensity methods can reach decision quality comparable to CASE at zero audit spend. The predeclared comparison of CASE with exact-propensity doubly robust estimation is $-0.0023$ with CI $[-0.0100,+0.0055]$ and is unresolved. Estimated propensities perform worse in this replay. Under distorted scoring every assignment-aware estimator still leaves the measurement component because reweighting the same recorded score does not change what that score measures.

\begin{table}[htbp]\centering\footnotesize\setlength{\tabcolsep}{5pt}
\caption{Target-value loss of log-only causal estimators under selective routing over $30$ partitions. The estimators use no randomized-audit budget but require the logging propensities. The
cellwise-perfect repair is an oracle and carries no audit price.}
\label{tab:causalbase}
\begin{tabular}{@{}lccc@{}}
\toprule
Arm & Audit \$ & A1--D0 & A1--D1\\
\midrule
Trust the log & $0$ & $0.0499$ & $0.0469$\\
Outcome regression & $0$ & $0.0360$ & $0.0370$\\
Stabilized IPW, exact propensities & $0$ & $0.0327$ & $0.0335$\\
Cross-fitted DR, exact propensities & $0$ & $0.0360$ & $0.0348$\\
Stabilized IPW, estimated propensities & $0$ & $0.0375$ & $0.0384$\\
Cross-fitted DR, estimated propensities & $0$ & $0.0435$ & $0.0456$\\
CASE, randomized audit & $2.39$ & $0.0338$ & $0.0338$\\
Cellwise-perfect assignment repair & --- & $0.0000$ & $0.0030$\\
\bottomrule
\end{tabular}
\end{table}

\subsection{When the remaining range determines the allocation}

Repairing selective routing moves the centre of the quality range, but the stopping rule only becomes useful when the width that remains is small enough to fit inside one decision region. We test this directly by widening the common evaluator band for each partition and serving budget until the adverse solve changes the incumbent allocation. On LLMRouterBench the median tolerated spread falls from $0.0456$ at $1.25\times$ the cheapest feasible spend to $0.0026$ at $8\times$. At $\eta=0.02$, $21$ of $30$ partitions are still determined at the tight budget and none are at the loose one. RouterBench shows a weaker gradient. A tight serving budget removes more rivals on price alone, so a wider set $\mathcal U$ can remain inside one allocation region. Table~\ref{tab:cert} reports three of the seven budgets swept and Table~\ref{tab:certfull} reports all of them. On LLMRouterBench the median margin falls monotonically from $1.25\times$ to $8\times$ and then rises at $12\times$, where the serving budget stops binding on $12$ of the $30$ partitions and the problem is close to unconstrained. RouterBench binds at every budget and shows no trend. The sweep holds the centre fixed with the band alone in $e_{kj}$, so Table~\ref{tab:cert} and this one measure sensitivity to $\eta$ rather than a settlement rate. The same table reports the margin over the model-level set of Assumption~\ref{as:modellevel}, a strict subset of the box of \eqref{eq:cset}. Certifying over it needs the $2^{K}$ vertices of $\{C_k\}$ rather than one adverse table, since the worst case moves with the rival. Margins rise by about a third on average, $180$ of the $420$ cells strictly improve and $240$ are unchanged, and the count with margin at least $0.02$ goes from $97$ to $121$. At the measured $\eta$ of $0.128$ it moves one cell, from one to two. The two sets coincide whenever the binding rival differs from the incumbent on a single workload, and separate only when it differs on several with conflicting demands on one model.

\begin{table}[htbp]\centering\scriptsize\setlength{\tabcolsep}{4pt}
\caption{\textbf{The full budget sweep, over the free box of \eqref{eq:cset} and over the model-level set of Assumption~\ref{as:modellevel}.} Median margin over $30$ partitions and counts with margin at least $0.02$, under $\mathcal U$ as written and under the row-structured set $\{C:C_{kj}=C_k\}$. The last column gives the partitions on which the serving budget still binds.}
\label{tab:certfull}
\begin{tabular}{@{}lrcccccl@{}}
\toprule
 & & \multicolumn{2}{c}{median margin} & \multicolumn{2}{c}{margin $\ge0.02$} & \\
\cmidrule(lr){3-4}\cmidrule(lr){5-6}
Corpus & $\Gamma$ / cheapest & box & model-level & box & model-level & budget binds\\
\midrule
LLMRouterBench & $1.25\times$ & $0.0456$ & $0.0536$ & $21/30$ & $21/30$ & $30/30$\\
LLMRouterBench & $1.5\times$ & $0.0180$ & $0.0233$ & $15/30$ & $17/30$ & $30/30$\\
LLMRouterBench & $2\times$ & $0.0179$ & $0.0209$ & $14/30$ & $17/30$ & $30/30$\\
LLMRouterBench & $3\times$ & $0.0122$ & $0.0129$ & $7/30$ & $12/30$ & $30/30$\\
LLMRouterBench & $5\times$ & $0.0103$ & $0.0132$ & $5/30$ & $9/30$ & $30/30$\\
LLMRouterBench & $8\times$ & $0.0026$ & $0.0037$ & $0/30$ & $3/30$ & $30/30$\\
LLMRouterBench & $12\times$ & $0.0045$ & $0.0054$ & $3/30$ & $3/30$ & $18/30$\\
\addlinespace[2pt]
RouterBench & $1.25\times$ & $0.0132$ & $0.0157$ & $10/30$ & $14/30$ & $30/30$\\
RouterBench & $1.5\times$ & $0.0044$ & $0.0070$ & $3/30$ & $4/30$ & $30/30$\\
RouterBench & $2\times$ & $0.0133$ & $0.0171$ & $9/30$ & $11/30$ & $30/30$\\
RouterBench & $3\times$ & $0.0057$ & $0.0063$ & $5/30$ & $5/30$ & $30/30$\\
RouterBench & $5\times$ & $0.0023$ & $0.0029$ & $0/30$ & $0/30$ & $30/30$\\
RouterBench & $8\times$ & $0.0037$ & $0.0047$ & $1/30$ & $1/30$ & $30/30$\\
RouterBench & $12\times$ & $0.0101$ & $0.0130$ & $4/30$ & $4/30$ & $30/30$\\
\bottomrule
\end{tabular}
\end{table}

Those thresholds are useful only if the empirical contrasts that motivate them are not artifacts of one resampling scheme. We therefore recompute the main comparisons over partitions, over requests, and over benchmark slices. Table~\ref{tab:intervals} shows the result. The benefit of randomized evidence under selective routing stays separated from zero under all three schemes. The value-scale measurement floor is weaker under the five-slice cluster bootstrap, and the cost of auditing under matched-random assignment also loses separation from zero there. The pre-specified comparison between the same-evaluator and target-outcome gates remains unresolved. These weaker contrasts change how strongly we describe particular loss differences, but they do not change the containment question for $\mathcal U$.

\begin{table}[htbp]\centering\scriptsize\setlength{\tabcolsep}{3.2pt}
\caption{$95\%$ intervals for the confirmatory contrasts under three resampling schemes.
R-EPIG$_{\mathrm{corr}}$ is a decision-information acquisition rule that reveals the top-two
contender in the most uncertain column, run on the selection-adjusted surface. Bold marks a slice-cluster interval that includes zero where the partition and request intervals exclude it. A dash marks a contrast without a request- or slice-level analogue.}
\label{tab:intervals}
\begin{tabular}{@{}lrccc@{}}
\toprule
Contrast, $35\%$ budget & Mean & Partition & Request & Slice clusters\\
\midrule
A1--D0, CASE $-$ trust & $-0.0161$ & $[-.0219,-.0103]$ & $[-.0190,-.0132]$ & $[-.0410,-.0083]$\\
A1--D0, CASE $-$ R-EPIG$_{\mathrm{corr}}$ & $+0.0003$ & $[-.0019,+.0025]$ & $[-.0007,+.0013]$ & $[-.0017,+.0007]$\\
A1--D1, CASE $-$ trust & $-0.0130$ & $[-.0179,-.0081]$ & $[-.0159,-.0103]$ & $[-.0293,-.0087]$\\
A1--D1, CASE $-$ pop.-operational oracle & $+0.0309$ & $[+.0271,+.0346]$ & $[+.0278,+.0338]$ & $[+.0193,+.0444]$\\
A1--D1, measurement floor $\Gmeas$ & $+0.0030$ & $[+.0017,+.0043]$ & $[+.0011,+.0047]$ & $\mathbf{[-.0002,+.0259]}$\\
A0--D0, CASE $-$ trust & $+0.0107$ & $[+.0067,+.0147]$ & $[+.0083,+.0132]$ & $\mathbf{[-.0013,+.0142]}$\\
\addlinespace[2pt]
A1--D1, false switch, same $-$ ref. channel & $-0.0111$ & $[-.0269,+.0047]$ & --- & ---\\
Stream mismatch, false switch, same $-$ ref. channel & $-0.0278$ & $[-.0514,-.0042]$ & --- & ---\\
\bottomrule
\end{tabular}
\end{table}

\subsection{Measuring the remaining width on real data}

The controlled replay tells us what the method should do when the two errors are manipulated separately. The more important question for the uncertainty set is whether a comparable width appears when the discrepancy between the recorded score and the designated outcome is observed rather than imposed. KuaiRand-Pure \citep{gao2022kuairand} provides that comparison on production traffic. We fix eight well-supported content groups and eight workloads, split users into disjoint log, audit, and scoring pools, form $M$ from the standard stream, form randomized readout pools from the audit stream, and use a held-out randomized scoring pool to fix $\muref_{kj}$ from the designated like outcome. The outcome used to score the final allocation therefore does not choose the cells that CASE audits.

Table~\ref{tab:kuairandfull} shows what happens. The selection effect is small relative to the scoring gap, so removing assignment error has limited room to improve the decision. Increasing $n_{\mathrm{read}}$ from $30$ to $100$ to $300$ changes CASE only from $0.00157$ to $0.00148$ to $0.00149$ in traffic-weighted target-value loss. When both streams are scored with the designated outcome instead of clicks, the click-specific floor disappears. Randomized evidence changes the routing component of the information about $\muref$, while the evaluator component remains as width in $\mathcal U$.

\begin{table}[htbp]\centering\scriptsize\setlength{\tabcolsep}{5pt}
\caption{The KuaiRand comparison in full with uniform randomized auditing added. Target-value loss is measured in units of the KuaiRand like indicator multiplied by $10^3$ at the $35\%$ evidence budget. Means and standard errors are over $30$ partitions. Under the target evaluator the
cellwise-perfect repair deploys the reference table itself, zero by construction, and the dash
marks the traffic vote on the randomized stream, which has no production traffic to read.}
\label{tab:kuairandfull}
\begin{tabular}{@{}lcccc@{}}
\toprule
& \multicolumn{2}{c}{evaluator = click (D1)} & \multicolumn{2}{c}{evaluator = target (D0)}\\
\cmidrule(lr){2-3}\cmidrule(l){4-5}
Deployed decision & A1 selective & A0 randomized & A1 selective & A0 randomized\\
\midrule
Trust-the-log      & $1.76\pm0.14$ & $1.49\pm0.07$ & $1.76\pm0.10$ & $1.94\pm0.08$\\
Uniform audit      & $1.59\pm0.06$ & $1.59\pm0.06$ & $1.83\pm0.05$ & $1.80\pm0.07$\\
CASE               & $1.57\pm0.06$ & $1.60\pm0.06$ & $1.74\pm0.06$ & $1.72\pm0.06$\\
\midrule
Traffic vote       & $1.94\pm0.04$ & ---           & $1.94\pm0.04$ & ---\\
Cellwise-perfect repair & $2.01\pm0.02$ & $2.02\pm0.02$ & $0.00$ & $0.00$\\
\bottomrule
\end{tabular}
\end{table}

A natural response to scoring error is to use the same evaluator again on held-out requests before changing deployment. We test this because another sample can reduce noise even when it cannot change the construct being measured. Both a same-evaluator check and a target-outcome check reduce harmful switches relative to deploying the audited candidate without a gate. The pre-specified prediction that the same-evaluator check would leave more harmful switches is not supported. The paired difference is $-0.011$ with CI $[-0.027,+0.005]$. More importantly, the population comparison made by that evaluator still disagrees with the designated outcome on $4.6\%$ of contested workloads and on $8.3\%$ under the stronger distortion. Repeating the same scoring channel can therefore make the estimate more precise without establishing that it measures $\muref$.

The same problem appears with real automatic judges. On LMArena, \texttt{gpt-4o-mini} judges every battle in both answer orders with model identity hidden. Among $19{,}368$ parsable battles it agrees with the human vote on $0.430$, flips after swapping answer order on $0.335$, and prefers the longer answer at $+0.243$ where the human verdicts it is meant to represent do at $+0.130$ (Table~\ref{tab:meas}). On a shared $275$-battle subset, a locally hosted \texttt{Qwen2.5-7B} agrees on $0.300$ and flips on $0.461$, with a length preference of the opposite sign. A two-way decomposition on the supported LMArena grid attributes $53.4\%$ of the \texttt{gpt-4o-mini} error to model main effects, compared with a permutation-null median of $13.0\%$. Applying the construction of \S\ref{sec:kuairand} to the same grid, the evaluator gap runs from $-0.169$ to $+0.184$ across the $48$ models, and the median within-workload spread is $\eta=0.336$ on the single-order scores an operator logs, $0.330$ once answer order is averaged out. HelpSteer2 gives the same qualitative warning with correctness as the designated outcome. The judge agrees with human correctness on $0.717$ of $1{,}182$ prompts, yet conditioning on correctness still leaves systematic effects of verbosity and complexity, and the judge selects a different response than human correctness on $28.3\%$ of prompts. These measurements support carrying evaluator error as remaining width around $\muref$ rather than treating randomized re-evaluation as point recovery.

\begin{table}[htbp]\centering\scriptsize\setlength{\tabcolsep}{4.5pt}
\caption{\textbf{Randomized reruns do not establish measurement validity.} LMArena
\citep{chiang2024arena} and HelpSteer2 \citep{wang2024helpsteer2}. \emph{Agrees} is raw agreement, not
chance-corrected, position-flip the share of battles whose verdict changes when answers are swapped,
and \emph{Longer answer} the judge's partial length preference. HelpSteer2 is rated one response at a
time, so it has no position to swap and the dash marks what the corpus cannot define. Its agreement
compares the responses within each of the $1{,}182$ prompts holding more than one, and its length
preference the $1{,}160$ of those holding two of unequal length.}
\label{tab:meas}
\begin{tabular}{@{}llrrrr@{}}
\toprule
Evaluator & Corpus and target & Items & Agrees & Position-flip & Longer answer\\
\midrule
\texttt{gpt-4o-mini} & LMArena, human verdict & $19{,}368$ & $0.430$ & $0.335$ & $+0.243$\\
\midrule
\texttt{gpt-4o-mini} & HelpSteer2, human correctness & $2{,}400$ & $0.717$ & --- & $+0.085$\\
\bottomrule
\end{tabular}
\end{table}

\subsection{What better information about the quality table is worth}

The remaining width matters economically only if changing the information supplied to the optimizer changes a deployment that is valuable enough to justify measurement. We therefore keep the decision problem fixed and compare allocations formed from progressively better information about $\muref$. On LLMRouterBench, the raw log gives target quality $0.4798$, the selection-adjusted table raises it to $0.4924$, and the designated outcome raises it to $0.5255$. The corresponding deployment spend rises from $\$3{,}109$ to $\$3{,}850$ and then $\$4{,}386$ per million requests. Better information does not make the deployment cheaper. It lets the operator use more of the available budget on models that are genuinely better. The randomized audit itself costs about $\$2.39$ at the $35\%$ evidence budget on LLMRouterBench and $\$0.89$ on RouterBench.

Table~\ref{tab:econextra} puts that comparison beside a dollar-valued task set. On $150$ SWE-Lancer \citep{miserendino2025swelancer} managerial tasks, routing on the observable success estimate reaches total cost $\$70{,}625$, only about $\$1{,}000$ below the best single model. Routing on the realized per-task outcome reaches $\$35{,}625$. The roughly $\$35{,}000$ gap between those two routed decisions is the economic value of the information missing from the observable quality estimate in this calculation. The dominant opportunity is therefore better information, not a more elaborate optimizer applied to the same table. That block uses four models over the $150$ tasks, grouped into four payout bands that play the role of workloads. Each option is priced by its expected total cost, the attempt cost plus the released payout when the model fails, and a direct human fallback is allowed, so the comparison carries no serving budget and neither \eqref{eq:mckp} nor \eqref{eq:sat} applies. The observable arm routes on the per-model, per-band success rate, which is this setting's analogue of $\muref$, and the oracle routes on each task's realized outcome.

\begin{table}[htbp]\centering\scriptsize\setlength{\tabcolsep}{5pt}
\caption{Economic scale of better quality information. The LLMRouterBench block reports target quality and deployment cost per million requests under the budgeted rule. The SWE-Lancer block reports total expected cost over $150$ managerial tasks, where lower is better. Quality is not reported for SWE-Lancer because the human
fallback completes every task, making the comparison one of cost alone.}
\label{tab:econextra}
\begin{tabular}{@{}llcc@{}}
\toprule
Setting & Information used for the decision & Target quality & Deployment or total cost\\
\midrule
LLMRouterBench & raw production log & $0.4798\se{0.0041}$ & $\$3{,}109\se{186}$ per 1M\\
LLMRouterBench & selection-adjusted scores & $0.4924\se{0.0027}$ & $\$3{,}850\se{91}$ per 1M\\
LLMRouterBench & designated outcome & $0.5255\se{0.0021}$ & $\$4{,}386\se{39}$ per 1M\\
\addlinespace[2pt]
SWE-Lancer & best single model & --- & $\$71{,}625$ total\\
SWE-Lancer & observable success estimate & --- & $\$70{,}625$ total\\
SWE-Lancer & realized per-task outcome & --- & $\$35{,}625$ total\\
\bottomrule
\end{tabular}
\end{table}

\section{Relation to interval optimization}\label{app:related}

Optimization under interval data usually starts from a given uncertainty set and asks for a decision that is
best or safest against the values that set permits \citep{kouvelis1997robust,averbakh2001complexity,aissi2009minmax}.
The question here comes one step earlier. The set $\mathcal U$ is derived from the production log, randomized
re-evaluation, and the remaining measurement ambiguity. We ask whether the current allocation is already
invariant throughout that set. For the budgeted rule, Prop.~\ref{prop:robust} answers this with one adverse
re-solve of the same optimization problem. When the test fails, the disagreement between the two allocations
also identifies the cells where reducing uncertainty can change the verdict. This connection from measured
uncertainty to a stopping test and then to the next evidence purchase is what this paper adds to
interval optimization.


\newpage
\end{document}